\documentclass{article}

\usepackage{aclap}

\usepackage{epsfig}

\usepackage{lingmacros}

\usepackage{times}

\usepackage{program}

\usepackage{pnnamed}

\begin{document}

\title{\bf Centering in-the-Large:\\
Computing Referential Discourse Segments}

\author{\bf{Udo Hahn \& Michael Strube} \\
	{\small \raisebox{-0.25mm}{\epsfxsize=5ex\epsfbox{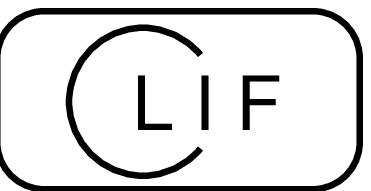}}}
		Computational Linguistics Research Group\\
	Freiburg University, Werthmannplatz 1\\
	D-79085 Freiburg, Germany\\
	\small{http://www.coling.uni-freiburg.de/}\\
}

\maketitle

\begin{abstract}
\noindent
We specify an algorithm that builds up a hierarchy of referential
discourse segments from local centering data. The spatial extension
and nesting of these discourse segments constrain the reachability of
potential antecedents of an anaphoric expression beyond the local
level of adjacent center pairs. Thus, the centering model is scaled up
to the level of the global referential structure of discourse. An
empirical evaluation of the algorithm is supplied.
\end{abstract}

\date{}

\section{Introduction}
\vspace{-1ex}
The centering model \cite{grosz95} has evolved as a major methodology
for computational discourse analysis. It provides simple, yet powerful
data structures, constraints and rules for the {\em local}\ coherence
of discourse. As far as anaphora resolution is concerned, e.g.,
the model requires to consider
those discourse entities as potential antecedents 
for anaphoric expressions in the current utterance $U_i$, 
which are available in the forward-looking centers
of the {\em immediately preceding} utterance $U_{i-1}$.
No constraints or rules are formulated, however, that account for
anaphoric relationships which spread out over non-adjacent utterances.
Hence, it is unclear how discourse elements which appear in utterances
preceding utterance $U_{i-1}$ are taken into consideration
as potential antecedents for anaphoric expressions in $U_i$.

The extension of the search space for antecedents is by no means a
trivial enterprise. A simple linear backward search of all preceding
centering structures, e.g., may not only turn out to establish illegal
references but also contradicts the cognitive principles underlying
the limited attention constraint \cite{walker96a}.  The solution we
propose starts from the observation that additional constraints on
valid antecedents are placed by the {\em global} discourse structure
previous utterances are embedded in.  We want to emphasize from the
beginning that our proposal considers only the {\em referential}
properties underlying the global discourse structure.
Accordingly, we define the {\em extension} of referential discourse
segments (over several utterances) and a {\em hierarchy} of
referential discourse segments (structuring the entire discourse).
\footnote{
Our notion of {\em referential} discourse segment should not be
confounded with the {\em intentional} one
originating from \textcite{grosz86}, for reasons discussed in Section
\ref{global}.
}
The algorithmic procedure we propose for creating and managing such
segments receives local centering data as input and
generates a sort of superimposed index structure by which the
reachability of potential antecedents, in particular those prior to
the immediately preceding utterance, is made explicit.
The adequacy of this definition is judged by the effects 
centered discourse segmentation has on
the validity of anaphora resolution (cf.\ Section \ref{empirical} for 
a discussion of evaluation results).

\section{Global Discourse Structure}\label{global}
\vspace{-1ex}
There have been only few attempts at dealing with the
recognition and incorporation of discourse structure beyond the level
of immediately adjacent utterances within the centering framework.
Two recent studies deal with this topic in order
to relate attentional and intentional structures on a larger scale
of global discourse coherence.
\textcite{passonneau96} proposes an algorithm for the generation of
referring expressions and \textcite{walker96b} integrates centering
into a cache model of attentional state.  Both studies, among other things,
deal with the supposition whether a correlation exists between
particular centering transitions (which were first introduced by
\textcite{brennan87}; cf.\ Table \ref{tab:trans}) and intention-based
discourse segments.  In particular, the role of {\sc shift}-type
transitions is examined from the perspective of whether they not only
indicate a shift of the topic between two immediately successive
utterances but also signal (intention-based) segment boundaries. The
data in both studies reveal that only a weak correlation between the
{\sc shift} transitions and segment boundaries can be observed.  This
finding precludes a reliable prediction of segment boundaries based on
the occurrence of {\sc shift}s and {\em vice versa}.  In order to
accommodate to these empirical results divergent solutions are proposed.  
\citeauthor{passonneau96} suggests that the centering data
structures need to be modified appropriately, while
\citeauthor{walker96b} concludes that the local centering data should
be left as they are and further be complemented by a cache mechanism.
She thus intends to extend the scope of centering in accordance with cognitively
plausible limits of the attentional span. Walker, finally, claims that the
content of the cache, rather than the intentional discourse segment
structure, determines the accessibility of discourse entities for
anaphora resolution.

\begin{table}[htb]
\centering
\begin{small}
\begin{tabular}{l|c|c}
& $C_b(U_n)$ = $C_b(U_{n-1})$ & $C_b(U_n)$ $\not=$ \\
& OR $C_b(U_{n-1})$ undef.\ & $C_b(U_{n-1})$ \\
\hline
$C_b(U_n)$ = & & \\
$C_p(U_n)$ & \raisebox{1.3ex}[-1.3ex]{{\sc continue (C)}} &
\raisebox{1.3ex}[-1.3ex]{{\sc smooth-shift (SS)}} \\
\hline
$C_b(U_n)$ $\not=$ & & \\
$C_p(U_n)$ & \raisebox{1.3ex}[-1.3ex]{{\sc retain (R)}} &
\raisebox{1.3ex}[-1.3ex]{{\sc rough-shift (RS)}} \\
\hline
\end{tabular}
\end{small}
\vspace{-1ex}
\caption{\label{tab:trans}Transition Types}
\vspace{-1ex}
\end{table}

As a working hypothesis, for the purposes of
anaphora resolution we subscribe to \citeauthor{walker96b}'s
model, in particular to that part which casts doubt on the
hypothesized dependency of the attentional from the intentional
structure of discourse \cite[p.180]{grosz86}.
We diverge from \textcite{walker96b}, however, in that we propose an
alternative to the caching mechanism, which we consider to be
methodologically more parsimonious and, at least, to be equally effective
(for an elaboration of this claim, cf.\ Section \ref{relat}).

The proposed extension of the centering model
builds on the methodological framework of
{\em functional centering} \cite{strube.acl96}.
This is an approach to centering 
in which issues such as thematicity or topicality are already inherent.  
Its linguistic foundations
relate the ranking of the {\em forward-looking centers}\ and the {\em
functional information structure}\ of the utterances, a notion
originally developed by \textcite{danes74a}.
\textcite{strube.acl96} use the centering data structures
to redefine \citeauthor{danes74a}'s trichotomy 
between {\em given information}, {\em theme} and {\em rheme}
in terms of the centering model.  
The $C_b(U_n)$, the most highly ranked element of
$C_f(U_{n-1})$ realized in $U_n$, corresponds to the element which
represents the {\em given}\ information. The {\em theme}\ of $U_n$ is
represented by the preferred center $C_p(U_n)$, the most highly ranked
element of $C_f(U_n)$. The {\em theme/rheme hierarchy}\ of $U_n$ 
corresponds to the ranking in the $C_f$s.  As a consequence,
utterances without any anaphoric expression do not have any {\em
given}\ elements and, therefore, no $C_b$. But independent of the use of
anaphoric expressions, each utterance must have a theme and a $C_f$ as
well.

The identification of the {\em preferred center}\ with the {\em
theme}\ implies that it
is of major relevance for determining the thematic progression of a text.
This is reflected in our reformulation of the two
types of thematic progression (TP)
which can be directly derived from centering data
(the third one requires to refer to conceptual generalization hierarchies
and is therefore beyond the scope of this paper, cf.\ \textcite{danes74a}
for the original statement):

\begin{enumerate}

\item {\em TP with a constant theme:}\ Successive utterances continuously share
the same $C_p$.

\item {\em TP with linear thematization of rhemes:}\ An element of the
$C_f(U_{i-1})$ which is not the $C_p(U_{i-1})$ appears in $U_i$
and becomes the $C_p(U_i)$ after the processing of this utterance.

\end{enumerate}

\begin{table}[htb]

\centering

\begin{tabular}{|lll|}

\hline

$C_f(U_{i-1}):$ & [ $c_1$, ..., $c_j$, ..., $c_s$ ] & \\

& \hspace{0.8em}$\downarrow$  &\\

$C_f(U_{i}):$ & [ $c_1$, ..., $c_k$, ..., $c_t$ ] & \\

\hline\hline

$C_f(U_{i-1}):$ & [ $c_1$, ..., $c_j$, ..., $c_s$ ] & $1 < j \leq
s$ \\

& \hspace{2em}$\swarrow$  & \\

$C_f(U_{i}):$ & [ $c_1$, ..., $c_k$, ..., $c_t$ ] & \\

\hline 

\end{tabular}

\vspace{-1ex}
\caption{Thematic Progression Patterns}
\label{tab:tp1}
\vspace{-1ex}
\end{table}

Table \ref{tab:tp1} visualizes the abstract
schemata of {\em TP patterns}. In
our example (cf.\ Table \ref{tab:center} in Section
\ref{sec:example}), $U_1$ to $U_3$ illustrate the {\em constant
theme}, while $U_7$ to $U_{10}$ illustrate the {\em linear
thematization of rhemes}. In the latter case, the theme changes in
each utterance, from {\em ``Handbuch'' (manual)}\ via {\em
``Inhaltsverzeichnis'' (table of contents)}\ to {\em ``Kapitel''
(chapter)}\ etc. Each of the new themes are introduced in the
immediately preceding utterance so that local coherence between these
utterances is established.

\textcite{danes74a} also allows for the combination and recursion of
these basic patterns; this way the global thematic coherence of a text
can be described by recurrence to these structural patterns.  These
principles allow for a major extension of the original centering
algorithm. Given a reformulation of the TP
constraints in centering terms, it is possible to determine
referential segment boundaries and to arrange these segments in a
nested, i.e., hierarchical manner on the basis of which reachability constraints
for antecedents can be formulated.  
According to the segmentation strategy of our
approach, the $C_p$ of the end point (i.e., the last utterance) of a
discourse segment provides the major theme of the whole segment, one which
is particularly salient for anaphoric reference relations.
Whenever a relevant new theme is established, however, it should
reside in its own discourse segment, either embedded or in parallel to
another one.  Anaphora resolution can then be performed {\em (a)} with the
forward-looking centers of the linearly immediately preceding utterance,  {\em (b)} with the forward-looking centers of the end point of the
hierarchically immediately reachable discourse segment, and {\em (c)} with the
preferred center of the end point of any hierarchically reachable
discourse segment (for a formalization of this constraint
, cf.\ Table \ref{tab:reach}).

\section{Computing Global Discourse Structure}
\vspace{-0,5ex}
Prior to a discussion of the algorithmic procedure for hypothesizing
discourse segments based on evidence from local centering data, we
will introduce its basic building blocks.  Let $x$ denote the
anaphoric expression under consideration, which occurs in utterance
$U_i$ associated with segment level $s$.  The function {\em
Resolved(x, s, $U_i$)}\ (cf.\ Table \ref{tab:resolv}) is evaluated in
order to determine the proper antecedent {\em ante}\ for $x$. It
consists of the evaluation of a reachability predicate for the
antecedent on which we will concentrate here, and of the evaluation of
the predicate {\em IsAnaphorFor}\ which contains the linguistic and
conceptual constraints imposed on a (pro)nominal anaphor 
({\em viz.} agreement, binding, and sortal constraints) or a textual ellipsis
\cite{hahn.ecai96}, not an issue in this paper. The predicate {\em
IsReachable}\ (cf.\ Table \ref{tab:reach}) requires {\em ante}\ to be
reachable from the utterance $U_i$ associated with the segment level
$s$.\footnote{
The $C_f$ lists in the functional centering model are {\em totally} ordered
\cite[p.272]{strube.acl96} and we here implicitly assume that they
are accessed in the total order given.
}
Reachability is thus made dependent on the segment structure $DS$
of the discourse as built up by the segmentation algorithm which is
specified in Table \ref{tab:algo}.  In Table \ref{tab:reach}, the
symbol ``$=_{str}$'' denotes string equality, {\bf N} the natural
numbers. We also introduce as a notational convention that a discourse
segment is identified by its index $s$ and its opening and closing
utterance, viz.\  $DS[s.beg]$ and $DS[s.end]$, respectively.
Hence, we may either identify an utterance $U_i$ by its linear text
index, $i$, or, if it is accessible, with respect to its hierarchical
discourse segment index, $s$ (e.g., cf.\ Table \ref{tab:center}
where $U_3 = U_{DS[1.end]}$ or $U_{13} = U_{DS[3.end]}$). The
discourse segment {\em index} is always identical to the currently
valid segment {\em level}, since the algorithm in Table \ref{tab:algo}
implements a stack behavior. Note also that we attach the discourse
segment index $s$ to center expressions, e.g., $C_b(s, U_i)$.

\begin{table}[htb]
\centering
\begin{small}
\fbox{
\begin{tabular}{l}
$
Resolved(x, s, U_i) := $ \\
$
\left\{ 
\begin{array}{lll}
ante & if & IsReachable(ante, s, U_i)  \\
& & \wedge \quad IsAnaphorFor (x, ante)\\
undef & else &
\end{array}
\right.
$
\end{tabular}
}
\end{small}
\vspace{-1ex}
\caption{Resolution of Anaphora}
\label{tab:resolv}
\vspace{-2ex}
\end{table}

\begin{table}[htb]
\centering
\begin{small}
\framebox[\columnwidth]{
\begin{tabular}{l}

$IsReachable(ante, s, U_i) $\\
$
\begin{array}{ll}
if & ante \in C_f(s, U_{i-1}) \\
else~if & ante \in C_f(s-1, U_{DS[s-1.end]}) \\
else~if & (\exists v \in {\bf N}: ante =_{str} C_p(v, U_{DS[v.end]}) \\
& \qquad \wedge~~~v < (s-1)) \\
& \wedge~~~(\neg \exists v' \in {\bf N}: ante =_{str} C_p(v',
U_{DS[v'.end]}) \\
& \qquad \wedge~~~v < v')
\end{array}
$
\end{tabular}
}
\end{small}
\vspace{-2ex}
\caption{Reachability of the Anaphoric Antecedent}
\label{tab:reach}
\vspace{-1ex}
\end{table}

Finally, the function {\em Lift(s, i)} (cf.\ Table \ref{tab:lift})
determines the appropriate discourse segment level, $s$, of an utterance $U_i$
(selected by its linear text index, $i$). {\em Lift}\ only applies to
structural configurations in the centering lists in which themes
continuously shift at three different consecutive segment
levels and associated preferred centers at least (cf.\
Table \ref{tab:tp1}, lower box, for the basic pattern).

\begin{table}[htb]
\centering
\begin{small}
\fbox{
\begin{tabular}{l}
$
Lift(s,i) := $ \\ 
$
\left\{ 
\begin{array}{l}
Lift(s-1, i-1) \quad if \\
\qquad s > 2 \wedge i > 3 \\ 
\qquad \wedge \quad C_p(s, U_{i-1}) \not= C_p(s-1, U_{i-2}) \\
\qquad \wedge \quad C_p(s-1, U_{i-2}) \not= C_p(s-2, U_{i-3}) \\
\qquad \wedge \quad C_p(s, U_{i-1}) \in C_f(s-1, U_{i-2}) \\
s \quad else
\end{array}
\right.
$
\end{tabular}
}
\end{small}
\vspace{-1ex}
\caption{Lifting to the Appropriate Discourse Segment}
\label{tab:lift}
\vspace{-1ex}
\end{table}

Whenever a discourse segment is created, its starting and closing
utterances are initialized to the current position in the
discourse. Its end point gets continuously incremented as the analysis
proceeds until this discourse segment $DS$ is {\em ultimately closed},
i.e., whenever another segment $DS'$ exists at the {\em
same}\ or a {\em hierarchically higher}\ level of embedding such that
the end point of $DS'$ exceeds that of the
end point of $DS$. Closed segments are inaccessible for the antecedent
search. In Table \ref{tab:center}, e.g., the first two discourse segments at
level 3 (ranging from $U_5$ to $U_5$ and $U_8$ to $U_{11}$)
are closed, while those at level 1 (ranging from $U_1$ to
$U_3$), level 2 (ranging from $U_4$ to $U_7$) and
level 3 (ranging from $U_{12}$ to $U_{13}$) are open.

The main algorithm (see Table \ref{tab:algo}) consists of three major
logical blocks ($s$ and $U_i$ denote the current discourse segment
level and utterance, respectively).

\begin{enumerate}
\vspace{-1ex}
\item {\bf Continue Current Segment.}\ The $C_p(s, U_{i-1})$ is
taken over for $U_i$. If $U_{i-1}$ and $U_i$ indicate the end of a
sequence in which a series of thematizations of rhemes have occurred,
all embedded segments are lifted by the function {\em Lift}\
to a higher level $s'$. As a result of lifting, the entire
sequence (including the final two utterances) forms a single
segment. This is
trivially true for cases of a constant theme.
\vspace{-1ex}
\item {\bf Close Embedded Segment(s).} 

\begin{enumerate}
\vspace{-1ex}
\item {\em Close the embedded segment(s) and continue another, already
existing segment:}\ If $U_i$ does not include any anaphoric expression
which is an element of the $C_f(s, U_{i-1})$, then match the antecedent
in the hierarchically reachable segments. Only the $C_p$ of the utterance
at the end point 
of any of these segments is considered
a potential antecedent.  Note that, as a side effect, 
hierarchically lower segments are
ultimately closed when a match at higher segment levels succeeds.
\vspace{-0,5ex}
\item {\em Close the embedded segment and open a new, parallel one:}\
If none of the anaphoric expressions under consideration co-specify
the $C_p(s-1, U_{[s-1.end]})$, then the entire $C_f$ at this
 segment level is checked for the given utterance. If an
antecedent matches, the segment which contains $U_{i-1}$ is
ultimately closed, since $U_i$ opens a parallel segment at the {\em
same} level of embedding. Subsequent anaphora checks exclude any of the
preceding parallel segments from the search for a valid antecedent
and just visit the currently open one.
\vspace{-0,5ex}
\item {\em Open new, embedded segment:}\ If there is no matching
antecedent in hierarchically reachable segments, then for utterance
$U_i$ a new, embedded segment is opened.
\vspace{-1ex}
\end{enumerate}

\item {\bf Open New, Embedded Segment.}\ If none of the above cases
applies, then for utterance $U_i$ a new, embedded segment is opened.
In the course of processing the following utterances, this decision
may be retracted by the function {\em Lift}. It serves as a kind of
``garbage collector'' for globally insignificant discourse segments
which, nevertheless, were reasonable from a local perspective for
reference resolution purposes. Hence, the centered discourse
segmentation procedure works in an incremental way and revises only
locally relevant, yet globally irrelevant segmentation decisions on
the fly.
\vspace{-1ex}
\end{enumerate}
\vspace{-1ex}

\begin{table}[htb]

\begin{small}

\begin{programbox}

s := 1 
i := 1 
DS[s.beg] := i 
DS[s.end] := i
\WHILE  \neg {\mbox ~| |end| |of| |text}
i := i + 1
{\cal R} := \left\{Resolved(x, s, U_i) \mid x \in U_i\right\}

\IF \exists r \in {\cal R}: r =_{str} C_p(s, U_{i-1}) \rcomment{\bf (1)}
\THEN s' := s
i' := i
DS[Lift(s',i').end] := i

\ELSE \IF \neg \exists r \in {\cal R}: r \in C_f(s, U_{i-1}) \rcomment{\bf (2a)}
\THEN found := |FALSE|
k := s
\WHILE \NOT found \AND (k > 1)
k := k-1

\IF \exists r \in {\cal R}: r =_{str} C_p(k, U_{[k.end]})
\THEN s := k
DS[s.end] := i
found := |TRUE|

\ELSE \IF k = s-1 \FI \rcomment{\bf (2b)}
\THEN \IF \exists r \in {\cal R}: r \in 
\hspace{0,5em}C_f(k, U_{[k.end]}) \FI
\THEN DS[s.beg] := i
DS[s.end] := i
found := |TRUE| \FI \FI \OD

\IF \neg found \rcomment{\bf (2c)}
\THEN s:= s+1 
DS[s.beg] := i
DS[s.end] := i \FI

\ELSE s:= s+1 \rcomment{\bf (3)}
DS[s.beg] := i
DS[s.end] := i \FI

\end{programbox}

\end{small}
\vspace{-1ex}
\caption{Algorithm for Centered Segmentation}
\label{tab:algo}
\vspace{-3ex}
\end{table}

\section{A Sample Text Segmentation}
\label{sec:example}
\begin{table*}[htb]

\framebox[\textwidth]{

\centering

\begin{tabular}{ll}

\begin{minipage}{\columnwidth}

\begin{small}

\enumsentence{
Brother HL-1260 
}

\enumsentence{
Ein Detail f{\"a}llt schon beim ersten Umgang mit dem gro{\ss}en
{\underline {\em Brother}}\ auf: \\
One particular -- is already noticed -- in the first approach to
-- the big {\underline {\em Brother}}:
}

\enumsentence{ Im Betrieb macht {\underline {\em er}}\ durch ein
kr{\"a}ftiges Arbeitsger{\"a}usch auf sich aufmerksam, das auch im
Stand-by-Modus noch gut vernehmbar ist. \\
In operation -- draws -- {\underline {\em it}} -- with a heavy
noise level -- attention to itself -- which -- also -- in the stand-by
mode -- is still well audible.
}

\enumsentence{F{\"u}r Standard-Installationen kommt man gut ohne
Handbuch aus. \\
As far as standard installations are concerned -- gets -- one -- well
-- by -- without any manual.
}

\enumsentence{Zwar erl{\"a}utert das d{\"u}nne {\underline {\em
Handb{\"u}chlein}}\ die Bedienung der {\em Hardware}\ anschaulich und gut
illustriert. \\
Admittedly, gives -- the thin {\underline {\em leaflet}}\ -- the
operation of the {\em hardware}\ -- a clear description of -- and -- well
illustrated.
}

\enumsentence{Die {\em Software-Seite}\ wurde im {\underline {\em
Handbuch}}\ dagegen stiefm{\"u}tterlich behandelt: \\
The {\em software part}\ -- was -- in the {\underline {\em manual}}\
-- however -- like a stepmother -- treated:
}

\enumsentence{bis auf eine karge {\em Seite}\ mit einem
Inhaltsverzeichnis zum {\em HP-Modus}\ sucht man vergebens weitere
Informationen. \\ except for one meagre {\em page}\ -- containing the
table of contents for the {\em HP mode}\ -- seeks -- one -- in vain --
for further information.
}

\end{small}

\end{minipage}

&

\begin{minipage}{\columnwidth}

\begin{small}

\enumsentence{Kein Wunder: unter dem {\underline {\em
Inhaltsverzeichnis}}\ steht der lapidare Hinweis, man m{\"o}ge sich
die Seiten dieses Kapitels doch bitte von Diskette ausdrucken --
Frechheit.  \\
No wonder: beneath the {\underline {\em table of contents}}\ -- one
finds the terse instruction, one should -- oneself -- the pages of
this section -- please -- from disk -- print out -- -- impertinence.
}

\enumsentence{Ohne diesen {\em Ausdruck}\ sucht man vergebens nach einem
Hinweis darauf, warum die {\em Auto-Continue-Funktion}\ in der
{\em PostScript-Emulation}\ nicht wirkt. \\
Without this {\em print-out}, looks -- one -- in vain -- for a hint -- why
-- the {\em auto-continue-function}\ -- in the {\em PostScript
emulation}\ -- does not work.
}

\enumsentence{Nach dem Einschalten zeigt das {\em LC-Display}\ an, da{\ss}
diese praktische {\underline {\em Hilfsfunktion}}\ nicht aktiv ist; \\
After switching on -- depicts -- the {\em LC display}\ -- that -- this practical
{\underline {\em help function}}\ -- not active -- is;
}

\enumsentence{{\underline {\em sie}}\ {\"u}berwacht den
Dateientransfer vom Computer.  \\
{\underline {\em it}}\ monitors the file transfer from the computer.
}

\enumsentence{Viele der kleinen Macken verzeiht man dem {\underline
{\em HL-1260}}\, wenn man erste Ausdrucke in H{\"a}nden h{\"a}lt. \\
Many of the minor defects -- pardons -- one -- the {\underline {\em
HL-1260}}, when -- one -- the first print outs -- holds in [one's] hands.
}

\enumsentence{Gerasterte Graufl{\"a}chen erzeugt der {\underline {\em
Brother}}\ sehr homogen \ldots  \\
Raster-mode grey-scale areas -- generates -- the {\underline {\em
Brother}}\ -- very homogeneously \ldots
}

\end{small}

\end{minipage}

\\

\end{tabular}

}

\vspace{-1ex}
\caption{Sample Text}
\label{tab:text}
\vspace{-2ex}
\end{table*}
\noindent
The text with respect to which we demonstrate the working
of the algorithm (see Table \ref{tab:text}) is taken from a German computer
magazine ({\em c't}, 1995, No.4, p.209). For ease of presentation
the text is somewhat shortened.
Since the method for computing levels of discourse segments
depends heavily on different kinds of anaphoric expressions, (pro)nominal
anaphors and textual ellipses are marked by italics, and the (pro)nominal
anaphors are underlined, in addition. In order to convey the influence
of the German word order we provide a rough phrase-to-phrase
translation of the entire text.

The centered segmentation analysis of the sample text is given in Table
\ref{tab:center}. The first column shows the linear text index of each
utterance. The second column contains the centering data as
computed by functional centering \cite{strube.acl96}. The
first element of the $C_f$, the {\em preferred center}, $C_p$, is
marked by bold font. The third column lists the centering
transitions which are derived from the $C_b/C_f$ data of immediately
successive utterances (cf.\ Table \ref{tab:trans} for the definitions).
The fourth column depicts the levels of discourse segments which
are computed by the algorithm in Table \ref{tab:algo}. 
Horizontal lines indicate the beginning of a segment 
(in the algorithm, this corresponds to
a value assignment to $DS[s.beg]$). 
Vertical lines show the extension of a segment (its
end is fixed by an assignment to $DS[s.end]$). The fifth column
indicates which block of the algorithm applies to the current
utterance (cf.\ the right margin in Table \ref{tab:algo}).

\begin{table*}[htb]

\centering

\begin{small}

\begin{tabular}{|c|ll|c|lllll|c|}

\hline

& & & &  \multicolumn{5}{|c|}{Levels of Discourse Segments} & \\
\raisebox{1.3ex}[-1.3ex]{$U_i$} & \multicolumn{2}{|c|}{\raisebox{1.3ex}[-1.3ex]{Centering Data}} & \raisebox{1.3ex}[-1.3ex]{Trans.} & 1 & 2 & 3 & 4
& 5 & \raisebox{1.3ex}[-1.3ex]{Block} \\

\hline

(1) & {\bf Cb:} & -- &  --- &
\rule[-1mm]{0.3mm}{3mm}\rule[1.7mm]{5mm}{0.3mm} &&&& &\\
& {\bf Cf:} & [{\bf 1260}] & &
\rule[-1mm]{0.3mm}{4mm} &&&&  & \\
\cline{1-4}

(2) & {\bf Cb:} & 1260 & C &
\rule[-1mm]{0.3mm}{4mm} &&&& & {\bf 1} \\
& {\bf Cf:} & [{\bf 1260}, Umgang, Detail] & &
\rule[-1mm]{0.3mm}{4mm} &&&& & \\
\cline{1-4}

(3) & {\bf Cb:} & 1260 & C &
\rule[-1mm]{0.3mm}{4mm} &&&& & {\bf 1} \\
& {\bf Cf:} & [{\bf 1260}, Betrieb, Arbeitsger{\"a}usch, Stand-by-Modus] & &
\rule[-1mm]{0.3mm}{4mm} &&&& & \\
\cline{1-4}

(4) & {\bf Cb:} & -- & --- & &
\rule[-1mm]{0.3mm}{3mm}\rule[1.7mm]{5mm}{0.3mm} &&& & {\bf 2c} \\
& {\bf Cf:} & [{\bf Standard-Installation}, Handbuch] & & &
\rule[-1mm]{0.3mm}{4mm} &&& & \\
\cline{1-4}

(5) & {\bf Cb:} & Handbuch & C & &
\rule[-1mm]{0.3mm}{4mm} &
\rule[-1mm]{0.3mm}{3mm}\rule[1.7mm]{5mm}{0.3mm} && & {\bf 3} \\
& {\bf Cf:} & [{\bf Handbuch}, 1260, Hardware, Bedienung] & & &
\rule[-1mm]{0.3mm}{4mm} &
\rule[-1mm]{0.3mm}{4mm} && & \\
\cline{1-4}

(6) & {\bf Cb:} & Handbuch & C & &
\rule[-1mm]{0.3mm}{4mm} &&& & {\bf 1, Lift} \\
& {\bf Cf:} & [{\bf Handbuch}, 1260, Software] & & &
\rule[-1mm]{0.3mm}{4mm} &&& & \\
\cline{1-4}

(7) & {\bf Cb:} & Handbuch & C & &
\rule[-1mm]{0.3mm}{4mm} &&& & {\bf 1} \\
& {\bf Cf:} & [{\bf Handbuch}, Seite, 1260, HP-Modus, & & &
\rule[-1mm]{0.3mm}{4mm} &&& & \\
& & Inhaltsverzeichnis, Informationen] & & & 
\rule[-1mm]{0.3mm}{4mm} &&& & \\
\cline{1-4}

(8) & {\bf Cb:} & Inhaltsverzeichnis & SS & &&
\rule[-1mm]{0.3mm}{3mm}\rule[1.7mm]{5mm}{0.3mm} && & {\bf 3} \\
& {\bf Cf:} & [{\bf Inhaltsverzeichnis}, Hinweis, Seiten, Kapitel, & & & &
\rule[-1mm]{0.3mm}{4mm} && & \\
& & Diskette, Frechheit] & & & &
\rule[-1mm]{0.3mm}{4mm} && & \\
\cline{1-4}

(9) & {\bf Cb:} & Kapitel & SS & &&
\rule[-1mm]{0.3mm}{4mm} &
\rule[-1mm]{0.3mm}{3mm}\rule[1.7mm]{5mm}{0.3mm} & & {\bf 3} \\
& {\bf Cf:} & [{\bf Kapitel}, Ausdruck, Hinweis, 1260, & & & &
\rule[-1mm]{0.3mm}{4mm} &
\rule[-1mm]{0.3mm}{4mm} & & \\
& & Auto-Continue-Funktion, PostScript-Emulation]& & & &
\rule[-1mm]{0.3mm}{4mm} &
\rule[-1mm]{0.3mm}{4mm} & & \\
\cline{1-4}

(10) & {\bf Cb:} & 1260 & RS & &&
\rule[-1mm]{0.3mm}{4mm} &&
\rule[-1mm]{0.3mm}{3mm}\rule[1.7mm]{5mm}{0.3mm}  & {\bf 3} \\
& {\bf Cf:} & [{\bf Auto-Continue-Funktion}, 1260, LC-Display] & & & &
\rule[-1mm]{0.3mm}{4mm} &&
\rule[-1mm]{0.3mm}{4mm}  & \\
\cline{1-4}

(11) & {\bf Cb:} & Auto-Continue-Funktion & SS & &&
\rule[-1mm]{0.3mm}{4mm} && & {\bf 1, Lift} \\
& {\bf Cf:} & [{\bf Auto-Continue-Funktion}, Dateien-Transfer, & & 
&&
\rule[-1mm]{0.3mm}{4mm} && & \\
& & Computer] & & & &
\rule[-1mm]{0.3mm}{4mm} && & \\
\cline{1-4}

(12) & {\bf Cb:} & -- & --- & &&
\rule[-1mm]{0.3mm}{3mm}\rule[1.7mm]{5mm}{0.3mm} && & {\bf 2b} \\
& {\bf Cf:} & [{\bf 1260}, Macken, Ausdruck] & & & &
\rule[-1mm]{0.3mm}{4mm} && & \\
\cline{1-4}

(13) & {\bf Cb:} & 1260 & C & &&
\rule[-1mm]{0.3mm}{4mm} && & {\bf 1} \\
& {\bf Cf:} & [{\bf 1260}, Graufl{\"a}chen] & & & &
\rule[-1mm]{0.3mm}{4mm} && & \\

\hline

\end{tabular}

\end{small}

\vspace{-1ex}
\caption{Sample of a Centered Text Segmentation Analysis}
\label{tab:center}
\vspace{-3ex}
\end{table*}

The computation starts at $U_1$, the headline. 
The $C_f(U_1)$ is set to {\em ``1260''}\ which is meant as
an abbreviation of {\em ``Brother HL-1260''}. 
Upon initialization, the beginning as well as the ending of the
initial discourse segment are both set to ``1''. 
$U_2$ and $U_3$ simply continue this
segment (block (1) of the algorithm), so {\em Lift}\ does not
apply. The $C_p$ is set to {\em ``1260''}\ in all utterances of this
segment. 
Since $U_4$ does neither contain any anaphoric expression which
co-specifies the $C_p(1, U_3)$ (block (1))
nor any other element of the $C_f(1, U_3)$ (block (2a)),
and as there is no hierarchically preceding segment,
block (2c) applies. The segment counter {\em s}\
is incremented and a new segment at level 2 is opened, setting the beginning
and the ending to ``4''.  The phrase {\em
``das d{\"u}nne Handb{\"u}chlein'' (the thin leaflet)}\ in $U_5$
does not co-specify the $C_p(2, U_4)$ but co-specifies an element of the $C_f(2,
U_4)$ instead ({\em viz.} {\em ``Handbuch'' (manual)}). 
Hence, 
block (3) of the algorithm applies, leading to the
creation of a new segment at level 3. 
The anaphor {\em ``Handbuch'' (manual)}\ in $U_6$
co-specifies the $C_p(3, U_5)$.  Hence block (1) applies (the
occurrence of {\em ``1260''}\ in $C_f(U_5)$ is due to the assumptions
specified by \textcite{strube.acl96}). 
Given this configuration, the function {\em Lift}\ 
lifts the embedded segment one level, so the
segment which ended with $U_4$ is now continued up to $U_6$ at level
2.  As a consequence, the centering data of $U_5$ are excluded from
further consideration as far as the co-specification by any subsequent
anaphoric expression is concerned.  $U_7$ simply continues the same
segment, since the textual ellipsis {\em ``Seite'' (page)}\ refers to
{\em ``Handbuch'' (manual)}. 
The utterances $U_8$ to $U_{10}$ exhibit a typical
 thematization-of-the-rhemes pattern which is quite common 
for the detailed description of objects. (Note the sequence of {\sc shift}
transitions.) Hence, block (3) of the algorithm applies to each of the
utterances and, correspondingly, new segments at the levels 3 to 5 are created. 
This behavior breaks down at the occurrence of the
anaphoric expression {\em ``sie'' (it)} in $U_{11}$ which co-specifies
the $C_p(5, U_{10})$, viz.\ {\em ``auto-continue
function''}, denoted by another anaphoric expression, namely
{\em ``Hilfsfunktion'' (help function)}\ in $U_{10}$. Hence, block (1)
applies.  \enlargethispage{\baselineskip}
The evaluation of {\em Lift}\ succeeds 
with respect to two levels of embedding. As a result, the whole
sequence is lifted up to level 3 and continues this segment which started
at the discourse element {\em ``Inhaltsverzeichnis'' (list of
contents)}. As a result of applying {\em Lift}, the whole sequence is
captured in one segment. $U_{12}$ does not contain any anaphoric
expression which co-specifies an element of the $C_f(3, U_{11})$, hence
block (2) of the algorithm applies. The anaphor {\em ``HL-1260''}\ does
not co-specify the $C_p$ of the utterance which represents the end of
the hierarchically preceding discourse segment ($U_7$), but it
co-specifies an element of the $C_f(2, U_7)$. The immediately preceding
segment is ultimately closed and a parallel segment is opened at $U_{12}$
(cf.\ block (2b)). Note also that the
algorithm does not check the $C_f(3, U_{10})$ despite the fact that it
contains the antecedent of {\em ``1260''}. However, the occurrences of
{\em ``1260''}\ in the $C_f$s of $U_9$ and $U_{10}$ are mediated by
textual ellipses. If these utterances contained the expression {\em ``1260''}\
itself, the algorithm would have built a different discourse structure and,
therefore, {\em ``1260''}\ in $U_{10}$ were reachable for the
anaphor in $U_{12}$. Segment 3, finally, is continued by $U_{13}$.

\section{Empirical Evaluation}\label{empirical}
In this section, we present some empirical data concerning the
centered segmentation algorithm.  Our study was based on the
analysis of twelve texts from the information technology domain (IT),
of one text from a German news magazine (Spiegel)
\footnote{Japan -- Der Neue der alten Garde. In {\em Der Spiegel},
Nr.\ 3, 1996.
}, and of two literary texts
\footnote{The first two chapters of a short story by the German writer
Heiner M{\"u}ller (Liebesgeschichte. In Heiner M{\"u}ller. {\em
Geschichten aus der Produktion 2}. Berlin: Rotbuch Verlag, 1974,
pp.57-63) and the first chapter of a novel by Uwe Johnson ({\em Zwei
Ansichten}. Frankfurt/Main: Suhrkamp Verlag, 1965.)
} (Lit). 
Table \ref{tab:testset} summarizes the total numbers of
anaphors, textual ellipses, utterances, and words in the test set.

\begin{table}[htb]
\centering

\begin{small} 

\begin{tabular}{|l|ccc|c|}

\hline

& IT & Spiegel & Lit & $\Sigma$ \\

\hline
anaphors & 197 & 101  & 198 & 496 \\

ellipses & 195 & 22 & 23 & 240 \\

utterances & 336 & 84 & 127 & 547 \\

words & 5241 & 1468 & 1610 & 8319 \\

\hline 

\end{tabular}

\end{small}
\vspace{-1ex}
\caption{Test Set}
\label{tab:testset}
\vspace{-1ex}
\end{table}

Table \ref{tab:dist-ana} and Table \ref{tab:dist-ell} consider the
number of anaphoric and text-elliptical expressions, respectively, 
and the linear distance they have to their corresponding antecedents. Note that
common centering algorithms (e.g., the one by \textcite{brennan87}) are
specified only for the resolution of anaphors in $U_{i-1}$. They are
neither specified for anaphoric antecedents in $U_i$, not an issue here,
nor for anaphoric antecedents beyond $U_{i-1}$. In the test set, 139
anaphors (28\%) and 116 textual ellipses (48,3\%) 
fall out of the (intersentential) scope of those common algorithms.
So, the problem we consider is not a marginal one.

\begin{table}[htb]
\centering

\begin{small}

\begin{tabular}{|l|ccc|c|}

\hline

& IT & Spiegel & Lit & $\Sigma$ \\

\hline
$U_i$ & 10 & 7 & 32 & 49 \\

$U_{i-1}$ & 117 & 70 & 121 & 308 \\

$U_{i-2}$ & 28 & 14 & 24 & 66 \\

$U_{i-3}$ & 18 & 5 & 10 & 33 \\

$U_{i-4}$ & 6 & 1 & 5 & 12 \\

$U_{i-5}$ & 6 & 0 & 1 & 7 \\

$U_{i-6}$ to $U_{i-10}$ & 8 & 1 & 3 & 12 \\

$U_{i-11}$ to $U_{i-15}$ & 3 & 1 & 1 & 5 \\

$U_{i-15}$ to $U_{i-20}$ & 1 & 2 & 1 & 4 \\

\hline 

\end{tabular}

\end{small}

\vspace{-1ex}
\caption{Anaphoric Antecedent in Utterance $U_x$}
\label{tab:dist-ana}
\vspace{-1ex}
\end{table}

\begin{table}[htb]
\centering

\begin{small}

\begin{tabular}{|l|ccc|c|}

\hline

& IT & Spiegel & Lit & $\Sigma$ \\

\hline

$U_{i-1}$ & 94 & 15 & 15 & 124 \\

$U_{i-2}$ & 42 & 6 & 8 & 56 \\

$U_{i-3}$ & 16 & 0 & 0 & 16 \\

$U_{i-4}$ & 14 & 0 & 0 & 14 \\

$U_{i-5}$ & 8 & 0 & 0 & 8 \\

$U_{i-6}$ to $U_{i-10}$ & 14 & 1 & 0 & 15 \\

$U_{i-11}$ to $U_{i-15}$ & 7 & 0 & 0 & 7 \\

\hline 

\end{tabular}

\end{small}

\vspace{-1ex}
\caption{Elliptical Antecedent in Utterance $U_x$}
\label{tab:dist-ell}
\vspace{-3ex}
\end{table}

Table \ref{tab:dist-ana2} and Table \ref{tab:dist-ell2} give the
success rate of the centered segmentation algorithm for anaphors and
textual ellipses, respectively.  The numbers in these tables indicate
at which segment level anaphors and textual ellipses were correctly
resolved.  The category of {\em errors}\ covers erroneous analyses the
algorithm produces, while the one for {\em false positives}\ concerns
those resolution results where a referential expression was resolved
with the hierarchically most recent antecedent but not with the
linearly most recent (obviously, the targeted) one (both of them
denote the same discourse entity). The categories $C_f(s, U_{i-1})$ in
Tables \ref{tab:dist-ana2} and \ref{tab:dist-ell2} contain more
elements than the categories $U_{i-1}$ in Tables \ref{tab:dist-ana}
and \ref{tab:dist-ell}, respectively, due to the mediating property of
textual ellipses in functional centering \cite{strube.acl96}.

\begin{table}[htb]
\centering

\begin{small}

\begin{tabular}{|l|ccc|c|}

\hline

& IT & Spiegel & Lit & $\Sigma$ \\

\hline
$U_i$ & 10 & 7 & 32 & 49 \\

$C_f(s, U_{i-1})$ & 161 & 78 & 125 & 364 \\

$C_p(s-1, U_{DS[s-1.end]})$ & 14 & 9 & 24 & 47 \\

$C_f(s-1, U_{DS[s-1.end]})$ & 7 & 5 & 9 & 21 \\

$C_p(s-2, U_{DS[s-2.end]})$ & 1 & 0 & 1 & 2 \\

$C_p(s-3, U_{DS[s-3.end]})$ & 1 & 0 & 1 & 2 \\

$C_p(s-4, U_{DS[s-4.end]})$ & 0 & 0 & 1 & 1 \\

$C_p(s-5, U_{DS[s-5.end]})$ & 0 & 1 & 0 & 1 \\

\hline

errors & 3 & 1 & 5 & 9 \\

\hline

false positives & (1) & (3) & (7) & (11) \\

\hline 

\end{tabular}

\end{small}

\vspace{-1ex}
\caption{Anaphoric Antecedent in Center$_x$}
\label{tab:dist-ana2}
\vspace{-2ex}
\end{table}

\begin{table}[htb]
\vspace{-1ex}
\centering

\begin{small}

\begin{tabular}{|l|ccc|c|}

\hline

& IT & Spiegel & Lit & $\Sigma$ \\

\hline

$C_f(s, U_{i-1})$ & 156 & 18 & 17 & 191 \\

$C_p(s-1, U_{DS[s-1.end]})$ & 18 & 0 & 4 & 22 \\

$C_f(s-1, U_{DS[s-1.end]})$ & 10 & 1 & 2 & 13 \\

$C_p(s-2, U_{DS[s-2.end]})$ & 7 & 1 & 0 & 8 \\

$C_p(s-3, U_{DS[s-3.end]})$ & 3 & 0 & 0 & 3 \\

\hline

errors & 1 & 2 & 0 & 3 \\

\hline

false positives & (2) & (0) & (3) & (5) \\

\hline 

\end{tabular}

\end{small}

\vspace{-1ex}
\caption{Elliptical Antecedent in Center$_x$}
\label{tab:dist-ell2}
\vspace{-1ex}
\end{table}

The centered segmentation
algorithm reveals a pretty good performance. This is
to some extent implied by 
the structural patterns we find in expository texts,
{\em viz.} their single-theme property (e.g., {\em ``1260''}\ in
the sample text). 
In contrast, the literary texts in the test
exhibited a much more difficult internal structure which resembled the
multiple thread structure of dialogues discussed by \textcite{rose95}.
The good news is that the segmentation procedure we propose
is capable of dealing even with these more complicated structures.
While only one antecedent of a pronoun was not reachable given the
superimposed text structure, the remaining eight errors are
characterized by full definite noun phrases or proper names. 
The vast majority of these phenomena can be considered {\em
informationally redundant utterances}\ in the terminology of
\textcite{walker96a} for which we
currently have no solution at all.  It seems to us that these kinds of
phrases may override text-grammatical structures as evidenced by
referential discourse segments and, rather, trigger other kinds of
search strategies.

Though we fed the centered segmentation algorithm with rather long
texts (up to 84 utterances), the antecedents of only two anaphoric expressions
had to bridge a hierarchical distance of more than 3
levels. This coincides with our supposition that the overall structure
computed by the algorithm should be rather flat. We could not find
an embedding of more than seven levels.

\section{Related Work}
\label{relat}
There has always been an implicit relationship between the local
perspective of centering and the global view of focusing on discourse
structure (cf.\ the discussion in \textcite{grosz95}). However, work
establishing an explicit account of how both can be joined in a
computational model has not been done so far.  The efforts of
\textcite{sidner83}, e.g., have provided a variety of different
focus data structures to be used for reference resolution.  This
multiplicity and the on-going growth of the number of different
entities (cf.\ \textcite{suri94}) mirrors an increase in
explanatory constructs that we consider a methodological drawback to
this approach because they can hardly be kept control of.  Our model,
due to its hierarchical nature implements a stack behavior
that is also inherent to the above mentioned proposals. We refrain,
however, from establishing a new data type (even worse, different
types of stacks) that has to be managed on its own. There is no need
for extra computations to determine the ``segment focus'', since that
is implicitly given in the local centering data already available in our
model.

A recent attempt at introducing global discourse notions into the
centering framework considers the use of a cache model
\cite{walker96a}.  This introduces an additional data type with its
own management principles for data storage, retrieval and update.
While our proposal for centered discourse segmentation also requires a
data structure of its own, it is better integrated into centering than
the caching model, since the cells of segment structures simply
contain ``pointers'' that implement a direct link to the original
centering data. Hence, we avoid extra operations related to feeding
and updating the cache. The relation between our centered segmentation
algorithm and \citeauthor{walker96b}'s \shortcite{walker96b}
integration of centering into the cache model can be viewed from two
different angles. On the one hand, centered segmentation may be a part
of the cache model, since it provides an elaborate, non-linear
ordering of the elements within the cache.  Note, however, that our
model does not require any {\em pre}fixed size corresponding to the
limited attention constraint.
On the other hand, centered segmentation may replace the
cache model entirely, since both are competing models of the
attentional state. Centered segmentation has also the additional
advantage of restricting the search space of anaphoric antecedents to
those discourse entities actually referred to in the discourse, while
the cache model allows unrestricted retrieval in the main or long-term
memory.

Text segmentation procedures (more with an information retrieval
motivation, rather than being related to reference resolution tasks)
have also been proposed for a coarse-grained partitioning of texts
into contiguous, non-overlapping blocks and assigning content labels
to these blocks \cite{hearst94}. The methodological basis of these
studies are lexical cohesion indicators \cite{morris91} combined with
word-level co-occurrence statistics. Since the labelling is
one-dimensional, this approximates our use of preferred centers of
discourse segments. These studies, however, lack the fine-grained
information of the contents of $C_f$ lists also needed for proper
reference resolution.

Finally, many studies on discourse segmentation highlight the role of
cue words for signaling segment boundaries (cf., e.g., the discussion
in \textcite{passonneau93b}).  However useful this strategy might be,
we see the danger that such a surface-level description may actually
hide structural regularities at deeper levels of investigation
illustrated by access mechanisms for centering data at different
levels of discourse segmentation.

\section{Conclusions}
We have developed a proposal for extending the centering model to
incorporate the global referential structure of discourse for reference
resolution.  The hierarchy of discourse segments we compute realizes
certain constraints on the reachability of antecedents. Moreover, the
claim is made that the hierarchy of discourse segments implements an
intuitive notion of the limited attention constraint, as we avoid a
simplistic, cognitively implausible linear backward search for
potentional discourse referents.  Since we operate within a functional
framework, this study also presents one of the rare formal accounts of
thematic progression patterns for full-fledged texts which were
informally introduced by \textcite{danes74a}.

The model, nevertheless, still has several restrictions.  First, it
has been developed on the basis of a small corpus of written texts.
Though these cover diverse text sorts ({\em viz.} technical product
reviews, newspaper articles and literary narratives), we currently do
not account for spoken monologues as modelled, e.g., by
\textcite{passonneau93b} or even the intricacies of dyadic
conversations \textcite{rose95} deal with. Second, a thorough
integration of the referential and intentional description of
discourse segments still has to be worked out.

\newpage
\begin{small}
\paragraph{Acknowledgments.} We like to thank our colleagues in the CLIF
group for fruitful discussions and instant support, 
Joe Bush who polished the text as a native speaker, 
the three anonymous reviewers for their critical comments, and,
in particular, Bonnie Webber for supplying invaluable comments 
to an earlier draft of this
paper. Michael Strube is supported by a post-doctoral grant from DFG
(Str 545/1-1).
\vspace{-1ex}
\end{small}
\setlength{\baselineskip}{-0.5ex}

\smallbibliography{\small}

\bibliographystyle{pnnamedabbrv}

\bibliography{/home/strube/lit/lit}

\end{document}